\documentclass[conference]{IEEEtran}

\usepackage[cmex10]{amsmath}
\usepackage{amsthm}
\usepackage{amssymb}
\usepackage{mathrsfs}
\usepackage{graphicx}
\usepackage{float}
\usepackage{array}
\usepackage{epstopdf}
\usepackage{multirow}
\usepackage{cite}

\begin{document}

\title{Palm-GAN: Generating Realistic
Palmprint Images Using Total-Variation Regularized GAN}

\author{Shervin Minaee$^1$,  Mehdi Minaei$^2$, AmirAli Abdolrashidi$^3$  \\
$^1$New York University\\
$^2$Sama Technical College, Azad University\\ 
$^3$University of California, Riverside\\ \\
}

\maketitle

\begin{abstract}
Generating realistic palmprint (more generally biometric) images has always been an interesting and, at the same time, challenging problem.
Classical statistical models fail to generate realistic-looking palmprint images, as they are not powerful enough to capture the complicated texture representation of palmprint images.
In this work, we present a deep learning framework based on generative adversarial networks (GAN), which is able to generate realistic palmprint images.
To help the model learn more realistic images, we proposed to add a suitable regularization to the loss function, which imposes the line connectivity of generated palmprint images. This is very desirable for palmprints, as the principal lines in palm are usually connected. 
We apply this framework to a popular palmprint databases, and generate images which look very realistic, and similar to the samples in this database.
%This framework can be used to augment the size of biometric datasets, enabling training of larger recognition models.
Through experimental results, we show that the generated palmprint images look very realistic, have a good diversity, and are able to capture different parts of the prior distribution.
We also report the Frechet Inception distance (FID) of the proposed model, and show that our model is able to achieve really good quantitative performance in terms of FID score.
\end{abstract}

\IEEEpeerreviewmaketitle
%Revise

\section{Introduction}
\label{sec:Intro}
Palmprint is one of the most widely used biometrics in today's research \cite{palm_intro}, alongside face \cite{face_intro}, fingerprints \cite{finger_old1}, and iris \cite{iris_intro}, for applications such as security and forensics. 
It contains a rich set of features, which can be used for identification and authentication.
All physiological biometrics are unique to the individual possessing them, and thus generating one that can pass as authentic can prove extremely difficult.
Many approaches have been proposed for palmprint recognition using minutiae-based features, geometry-based features, transformed based features \cite{palm_1, palm_2, palm_3, palm_4, palm_5}, and more.
Also, with deep neural networks achieving outstanding results in various computer vision applications during recent years \cite{cnn1,cnn2,cnn3,cnn4,cnn5}, end-to-end models have been heavily used for different biometric recognition problems \cite{deep_biometric, deep_face1,deep_face2, deep_finger1, deep_finger2, deep_iris1, deep_palm1, deep_iris2}.

%All biometrics, including fingerprints, have unique patterns. Therefore, generating an iris image would not be easy at all. However, a new trend has begun in recent years using a variety of different methods. Wei et al \cite{iris_gen1} managed to generate iris images using patch-based sampling. Makthal \cite{iris_gen2} proposed to use Markov random fields (MRF) to generate synthetic iris images. Cui \cite{iris_gen3} first synthesizes iris images using principal component analysis (PCA) and then uses super-resolution to enhance them. Zuo \cite{iris_gen4} uses a ``model-based anatomy-based'' approach, which includes generating 3D fibers in a cylindrical shape and projecting them on a 2D field, followed by adding blur effects and eyelids.

In this work, we focus on a different problem from recognition, namely generating synthetic palmprint images, by learning a generator model over training images.
This is extremely useful for training deep models, which are known to be data-hungry, as it can enable us to increase the training size of those datasets by a large factor.
Our work is mainly built upon  generative adversarial networks (GAN) \cite{gan}, which provides a powerful framework for learning to generate  samples from a given distribution. GAN contains two main components: the generative network, which can learn a dataset's distribution and generate new samples drawn from that distribution; and the discriminative network, whose job is to separate samples that belong to a dataset from the generated ones. 
These two networks are trained together through a minimax-type optimization. %, learning to generate samples similar to real data, and at the same time trying to discriminate real samples from ``fake'', i.e. the ones generated with the generator.
Since their development, GANs have been used for various computer vision and pattern generation problems, such as image super-resolution \cite{sr-gan}, image-to-image translation \cite{pix2pix}, text-based image generation \cite{gan_text}, pedestrian walking pattern prediction for autonomous vehicles \cite{gan_walk}, and also fingerprint \cite{fingergan} and iris generation \cite{irisgan}.

\begin{figure}[t]
\begin{center}
   \includegraphics[width=0.9\linewidth]{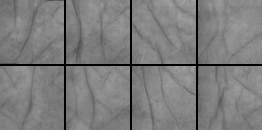}
\end{center}
   \caption{Eight sample generated palmprint images using the proposed framework. These images are generated with a model trained on PolyU palmprint database.}
\label{fig:gan_model}
\end{figure}

In this work, we propose a palmprint generation framework based on total variation regularized convolutional generative adversarial networks (TV-regularized DC-GAN), which is able to generate realistic palmprint images, which are hard to be distinguished from real ones. 
As the name suggests, we added a suitable regularization to the GAN loss function, imposing the line-connectivity in the generated palmprint images. 
This prior is very desirable for plamprint images, as most of the lines in palmprint are known to be connected.
This is achieved by adding the total variation \cite{TV} of the generated images to the loss function. 
Figure \ref{fig:gan_model} provide 8 sample images generated by our proposed model.

The structure of the remaining parts of this paper is as follows.
Section \ref{sec:Framework} provides the details of our proposed framework, and the architecture of the proposed generator, and discriminator networks.
Then, in Section \ref{sec:Evaluation}, we provide the experimental studies, and showcase some of the generated palmprint images with our proposed framework over different epochs, to see how the  images change over time.
Finally the paper is concluded in Section \ref{sec:Conclusion}.

\section{The Proposed Framework}
\label{sec:Framework}
In this section we provide the details of the proposed palmprint generation framework.
As mentioned eralier, our model is based on
generative adversarial networks, with a suitable regularization suited for palmprint generation.
First we shall give more technical details on GAN before diving into our proposed model.

\subsection{Generative Adversarial Network}
\begin{figure}[ht]
\begin{center}
   \includegraphics[page=1,width=0.98\linewidth]{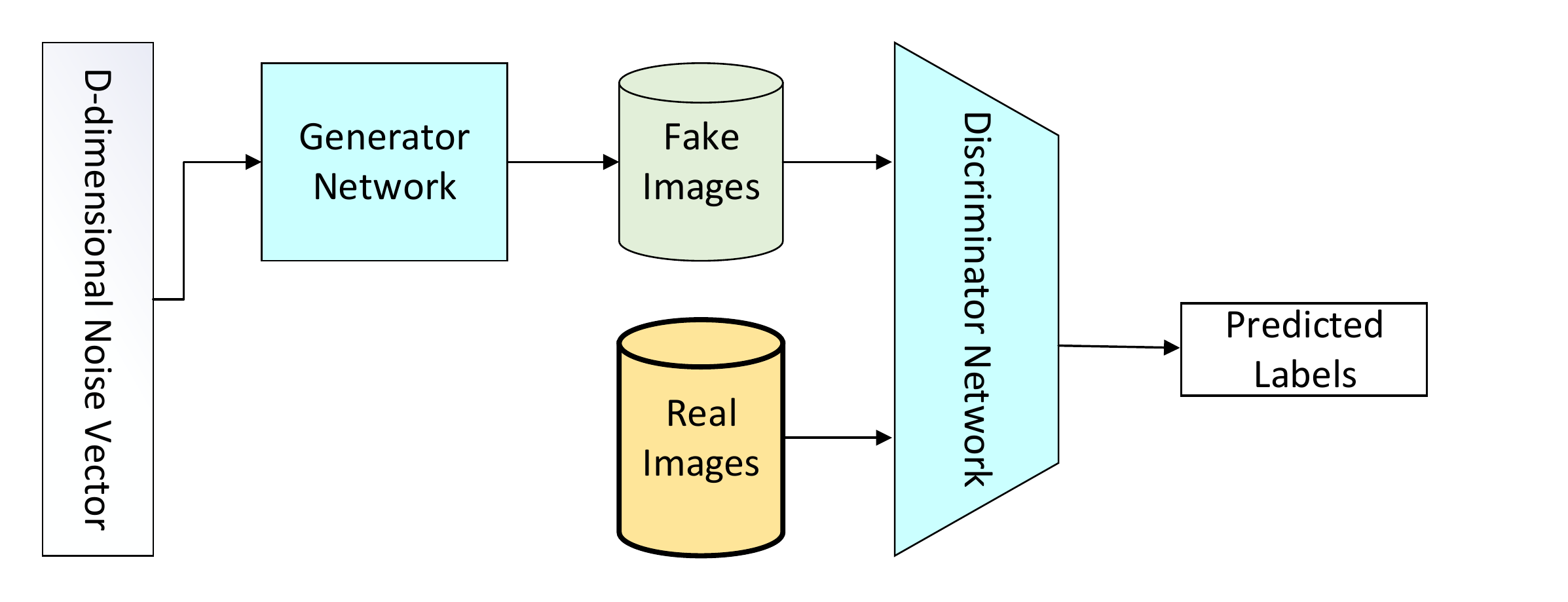}
\end{center}
   \caption{The general architecture of generative adversarial networks}
\label{fig:gen_arch}
\end{figure}

The generator model $G$ in GAN learns a mapping from noise $z$ (with a prior distribution) to a target distribution $y$, $G: z \rightarrow y$.
$G$ will attempt to generate samples which would be similar to the real samples provided during training. The discriminator network, $D$, on the other hand, will try to distinguish the real samples in the dataset from the others.
The general architecture of a GAN model is shown in Figure \ref{fig:gen_arch}.

The generator and discriminator can be trained jointly by minimizing a mini-max loss function.
The GAN's loss function can be written as:
\begin{equation*}
\begin{aligned}
& \mathcal{L}_{GAN}=  \mathbb{E}_{x \sim p_{data}(x)}[\text{log} D(x)]+ \mathbb{E}_{z \sim p_z(z)}[\text{log}(1-D(G(z)))]
\end{aligned}
\end{equation*}
In a way, a GAN resembles a minimax game between $D$ and $G$, where $G$ wants to maximize the error rate of $D$ (minimize $\mathcal{L}_{GAN}$), while $D$ wants to minimize its own error rate in distinguishing the source of its input samples (maximize $\mathcal{L}_{GAN}$).
After training this model, the trained generator model would be: 
\begin{equation}
\begin{aligned}
G^*= \text{arg} \ \text{min}_G \text{max}_D \ \mathcal{L}_{GAN}
\end{aligned}
\end{equation}
In practice, the above $\mathcal{L}_{GAN}$ may not provide enough gradient for $G$ to get well-trained, especially at the beginning of the training, where $D$ can easily detect fake samples from the real ones. 
One solution is to let the generator maximize $\mathbb{E}_{z \sim p_z(z)}[\text{log}(D(G(z)))]$ (instead of minimizing $\mathbb{E}_{z \sim p_z(z)}[\text{log}(1-D(G(z)))]$), which provides better gradient flow.

%During training, each time a batch of $m$ real examples, and $m$ noise samples are taken, and the generative and discriminative networks are updated alternatively.

The original paper on GAN proposed to use KL-divergence to measure the distance between distributions, which can be problematic if these distributions have non-overlapping support, in which KL divergence is zero, and is not a good representative of their distance. 
Since the beginning of GAN, there have been  works trying to improve or modify GAN in different aspects.
In \cite{dc-gan}, Radford et al. proposed a convolutional GAN model for generating images, which generates visually higher quality images than fully connected networks.
In \cite{con-gan}, Mirza et al. proposed a conditional GAN model, which can generate images conditioned on class labels. This enables one to generate samples with a specified label (such as specific digit when trained on MNIST dataset).
Also, in \cite{gan_text}, Zhang et al proposed a GAN-based high-resolution image generation network using a text description of the image.
%The work in \cite{was-gan} addresses this issue, by proposing a new loss function based on Wasserstein distance (also known as earth mover's distance).
In \cite{cycle-gan}, Zhu proposed an image to image translation model based on a cycle-consistent GAN model, which learns to map a given image distribution into a target domain (e.g. horse to zebra, day to night images, summer to winter images).
The idea of adversarial learning has also been applied to unsupervised setting (adversarial auto-encoders are a famous example of this category) \cite{adv_ae1, adv_ae2}, in which the latent representation are encouraged to come from a prior distribution.

\subsection{TV-Regularized GAN}
We use a deep convolutional GAN (DC-GAN) as the main core of our palmprint generation framework.
The palmprint images consist of lines, each forming a connected component.
Therefore, we want to extend our DC-GAN framework so that the model learns to generate palmprint images with their lines connected.
Although GAN  can learn this line connectivity from the dataset, we found that imposing a regularization term can greatly help with the connectivity of the generated palmprint images in practice. 
There are a few ways to impose such connectivity within the images \cite{TV, group-sparse}. We have used total variation (TV)  as its gradient update is relatively less complex during the backward pass.

TV penalizes the generated images with large variations among neighboring pixels, leading to more connected and smoother solutions \cite{TV1, TV2}. Total variation of a differentiable function $f$, defined on an interval $[a,b] \subset R$, has the following expression if $f'$ is Riemann-integrable:
\begin{equation}
\begin{aligned}
V_a^b= \int_{a}^{b} \| f'(x) \| dx
\end{aligned}
\end{equation}

Total variation of 1D discrete signals ($y=[y_1,  ..., y_N]$) is straightforward, and can be defined as:
\begin{equation}
\begin{aligned}
TV(y)=  \sum_{n=1}^{N-1} |y_{n+1}-y_{n}|
\end{aligned}
\end{equation}

\iffalse
where $D$ is a $(N-1)\times N$ matrix as below:
\begin{equation*}
D= 
\begin{bmatrix}
    1       & -1 & 0 & \dots & 0 \\
    0       & 1 & -1 & \dots & 0 \\
    \vdots & \vdots & \vdots & \ddots & \vdots \\
    0       & 0 & \dots &  1 & -1
\end{bmatrix}
\end{equation*}
\fi
For 2D signals ($Y=[y_{i,j}]$), we can use the isotropic or the anisotropic version of 2D total variation \cite{TV}. 
To simplify our optimization problem, we have used the anisotropic version of TV, which is defined as the sum of horizontal and vertical gradients at each pixel:
\begin{equation}
\begin{aligned}
TV(Y)=  \sum_{i,j} |y_{i+1,j}-y_{i,j}|+|y_{i,j+1}-y_{i,j}|
\end{aligned}
\end{equation}

Adding this 2D-TV regularization term to our framework will promote the connectivity of the produced images. The new loss function for our model would then be defined as:
\begin{equation}
\begin{aligned}
&  \mathcal{L}_{GAN-TV}=  \mathbb{E}_{x \sim p_{data}(x)}[\text{log} D(x)]+ \\ & \mathbb{E}_{z \sim p_z(z)}[\text{log}(1-D(G(z)))] + \lambda \ \text{TV}(G(z))
\end{aligned}
\end{equation}
Both the generator and the discriminator networks can be trained by optimizing this loss function, in an alternating fashion.

\subsection{Network Architecture}
\begin{figure}[ht]
\begin{center}
   \includegraphics[page=2,width=1\linewidth]{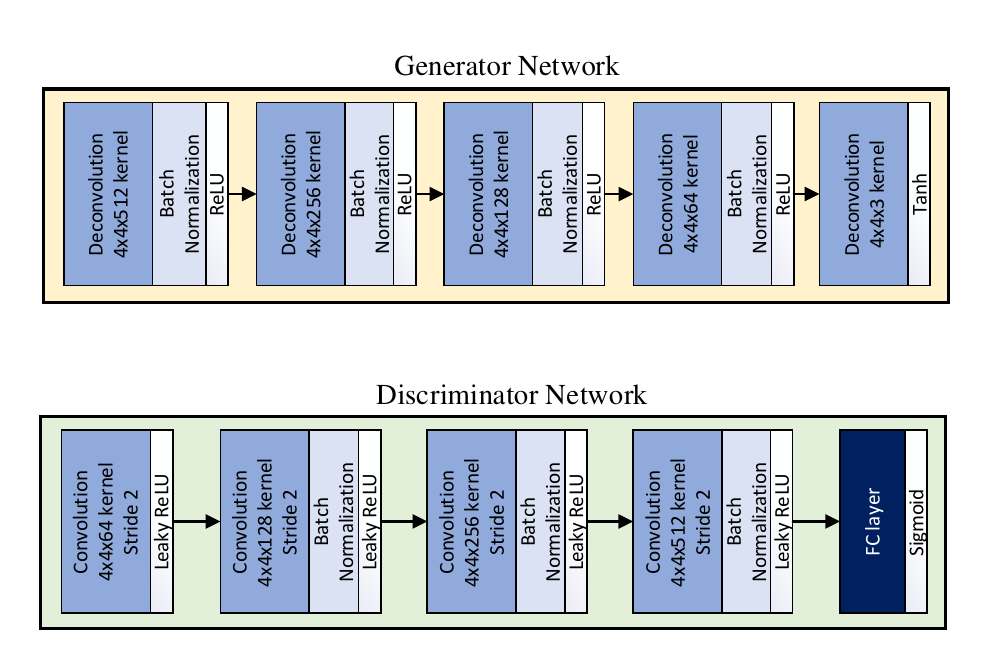}
\end{center}
   \caption{The architecture of our proposed generator (top) and discriminator model (bottom).}
\label{fig:our_model}
\end{figure}

In this part, we provide the details of the  architecture of both our generator and discriminator networks.
Both networks in our model are based on convolutional neural networks.
The generator model contains 5 fractionally-strided convolutional layers (deconvolution), followed by batch-normalization and non-linearity.
The discriminator model consists of 4 convolutional layers, followed by batch normalization and leaky ReLU as non-linearity, and a fully connected layer at the end.
The kernel size, and the number of feature maps of all layers are provided in  Figure \ref{fig:our_model}.
One can all use a similar network with fewer or more layers, depending on the applications. But we found that using a 5-layer network provides a good trade-off between the quality of the generated images and training time.

\section{Experimental Results}
\label{sec:Evaluation}
Before presenting the generated palmprint images by the proposed framework, let us first discuss the hyper-parameters used in our training procedure.
The proposed model is trained for 100 epochs on an Nvidia Tesla GPU.
The batch size is set to 40, number of workers to 4. 
ADAM optimizer is used to optimize the loss function, with a learning rate of 0.0002.
The input noise to the generator network is drawn from a 100-dimensional Gaussian distribution with zero mean, unit variance.
The details of the datasets used for our work are presented in the next section, followed by the experimental results.

\subsection{PolyU Plamprint Datasbe}
In this section, we provide a quick overview of the dataset used in this work, PolyU palmprint dataset \cite{polyu, polyu2}.
PolyU dataset has 6000
palmprints sampled from 500 persons. Each palmprint is taken under 4 different lights in two different days, is pre-processed and has a size of 128x128.
To make the training time faster, we resized all images to 64x64 resolution. 
%This is very helpful to speed up Generator network.

\begin{figure}[ht]
\begin{center}
   \includegraphics[width=0.9\linewidth]{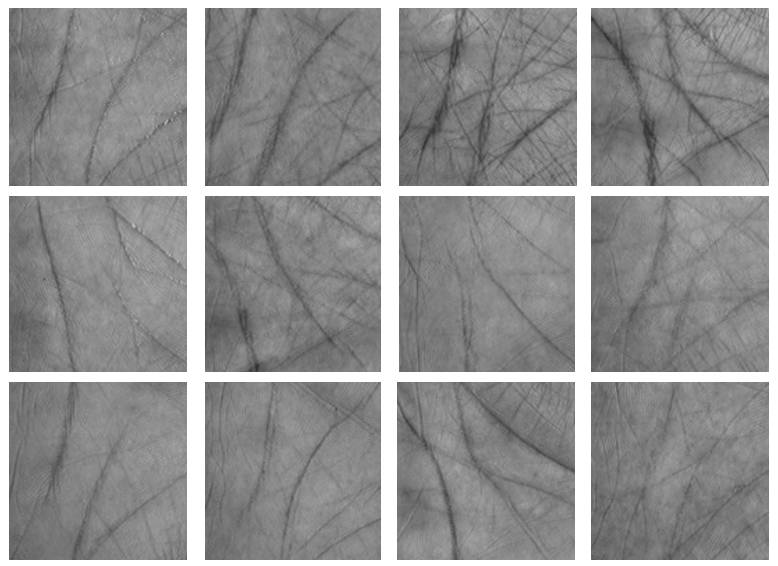}
\end{center}
   \caption{12 sample images from PolyU palmprint database.}
\label{fig:polyU12}
\end{figure}

Twelve sample images from this database are shown in Figure \ref{fig:polyU12}.

\subsection{Experimental Analysis and Generated Palmprint Images}
We will now present the experimental results of the proposed palmprint generator model trained on the PolyU palmprint dataset.
After training the model, one can generate different images by sampling latent representation from our prior Gaussian distribution, and feeding them to the generator network of the trained models.

\begin{figure}[ht]
\begin{center}
    \begin{tabular}{cc}
    \rotatebox{0}{0th Epoch} &
    \rotatebox{0}{20th Epoch} \\
    \includegraphics[width=0.45\linewidth]{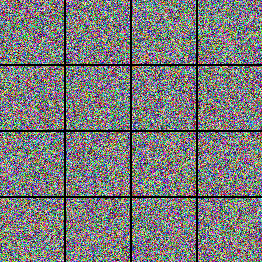}
    & \includegraphics[width=0.45\linewidth]{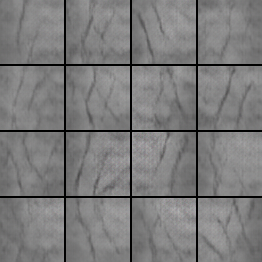}\\ 
    \rotatebox{0}{40th Epoch} &
    \rotatebox{0}{60th Epoch} \\
    \includegraphics[width=0.45\linewidth]{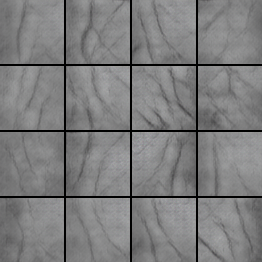}
    & \includegraphics[width=0.45\linewidth]{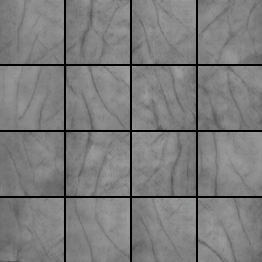}\\ 
    \rotatebox{0}{80th Epoch} &
    \rotatebox{0}{100th Epoch} \\
    \includegraphics[width=0.45\linewidth]{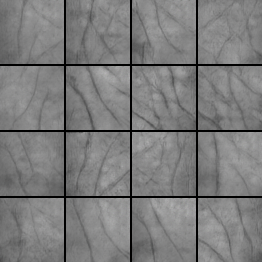}
    & \includegraphics[width=0.45\linewidth]{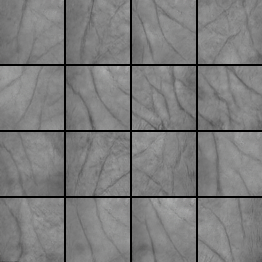}\\ 
    \end{tabular}
\end{center}
   \caption{Sixteen generated palmprint images using the same input noise, over different epochs.}
\label{fig:polyU16}
%\vspace{-3mm}
\end{figure}

In Figure \ref{fig:polyU16}, we present 16 generated palmprint images over different epochs (0th, 20th, 40th, 60th, 80th and 100th), to see the amount of diversity among the generated images (in terms of the position of the principal lines, the color of the palmprint, and their contrast).
As it can be seen, there is a good amount of diversity in all the aforementioned aspects across the 16 generated palmprints.

\begin{figure*}[t]
\begin{center}
   \includegraphics[width=0.9\linewidth]{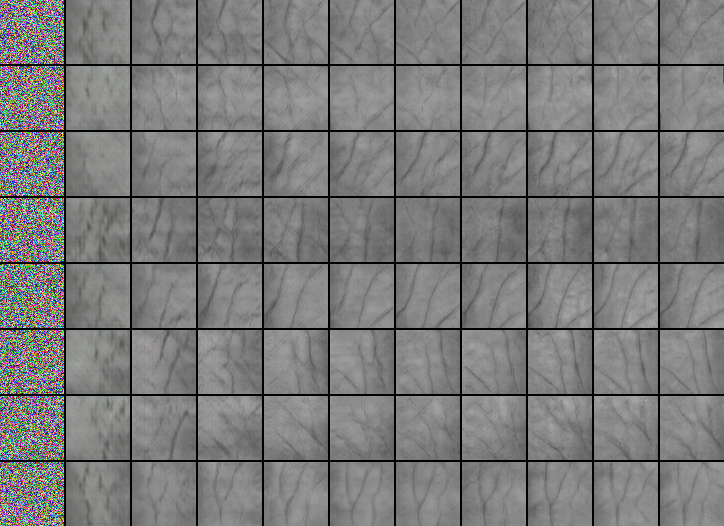}
\end{center}
   \caption{The generated palmprint images for 8 input latent vectors, over 100 epochs (on every 10 epochs), using the trained model on PolyU  palmprint database.}
\label{fig:polyU_10epochs}
\end{figure*}

In Figure \ref{fig:polyU_10epochs}, we show a similar results for four samples images, using the trained model on PolyU palmprint dataset, in every 10 epochs, to see the incremental improvement of the model.
As we can see the generative model gets more and more powerful over time, and the generated palmprint images gets more realistic over after more epochs.

\begin{figure}[tbh]
\begin{center}
   \includegraphics[width=0.95\linewidth, trim={0 1cm 0 0}]{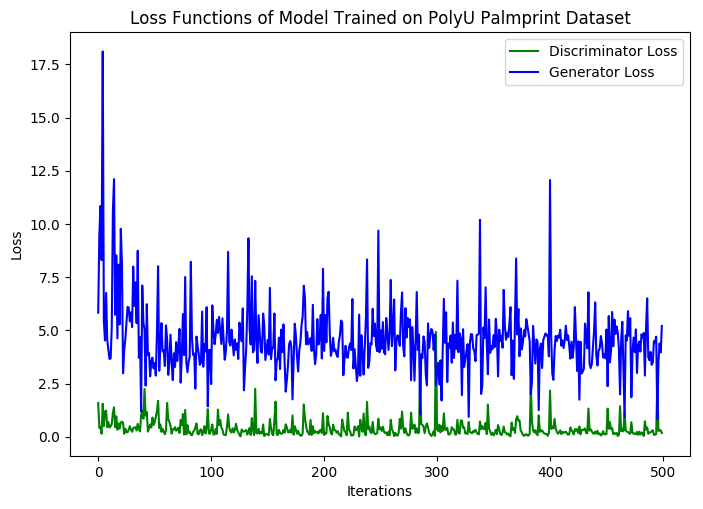}
\end{center}
\vspace{-2mm}
   \caption{The discriminator and generator loss functions over different iterations for the model trained on the PolyU palmprint dataset.}
   \vspace{-6mm}
\label{fig:polyU_loss}
\end{figure}

We also present the discriminator and generator loss functions for the trained Palm-GAN models on PolyU database, in Figure \ref{fig:polyU_loss}.
As it can be seen, the generator loss has a decreasing trend over time, while for discriminator loss we can observe such trend.

\subsection{Numerical Results}
To  measure  the quality  and  diversity of the model numerically,  we computed the Frechet Inception Distance (FID) \cite{FID} on the generated palmprint images by our model.
FID is an extension of Inception Score (IS) \cite{IS}, which was previously proposed for assessing the quality of generated images by GANs and other generative models.
FID compares the statistics of generated samples to real samples, using the Frechet distance between two multivariate Gaussians, defined as below:
\begin{equation}
\begin{aligned}
& \textbf{FID}=  \|\mu_r - \mu_g \|^2 + \textbf{Tr} (\Sigma_r + \Sigma_g - 2(\Sigma_r \Sigma_g)^{(1/2)} )
\end{aligned}
\end{equation}
where $X_r \sim N(\mu_r, \Sigma_r)$ and $X_g \sim N(\mu_g, \Sigma_g)$ are the 2048-dimensional activations of the Inception-v3 pool3 layer for real and generated samples respectively.

The FID scores by our proposed Palm-GAN model is shown in Table \ref{TblComp}.
As we can see, the model achieves relatively low FID score, comparable with some of state-of-the-arts generative models on public image datasets.

\begin{table}[ht]
\centering
  \caption{The Frechet Inception Distance for the trained Palm-GAN model}
  \centering
\begin{tabular}{|c|c|}
\hline
Model  & Frechet Inception Distance \\
\hline
TV-regularized DC-GAN  &   46.3 \\
\hline
\end{tabular}
\label{TblComp}
\end{table}

\section{Conclusion}
\label{sec:Conclusion}
In this work we propose a model for palmprint image generation, using a convolutional generative adversarial network.
We also add a suitable regularization term (the total variation of the output) to the loss function, to impose the connectivity of the generated palmprint images.
This is highly desirable for palmprint images, as the principal lines (and other lines) is palmprint usually form a connected component. 
Both the generator and discriminator networks of our model contain 5 layers, making it feasible to train them even on a laptop without GPU.
These models are trained on a popular palmprint dataset, (PolyU multi-spectral palmprint).
The experimental results advocate that the synthetic palmprint images look reasonably realistic and it is hard to distinguish them from the original palmprints.
As a future work,  we plan to train a coditional palmprint image generation model (using conditional-GAN), which is able to generate image samples for a given identity.
This would be very helpful for enlarging datasets for palmprint recognition task.

\section*{Acknowledgment}
We would like to thank the biometric research group at PolyU Hong Kong for providing the palmprint dataset  used in this work.
We would also like to thank Ian Goodfellow for his comments and suggestions regarding the analysis of the performance of this model.

\end{document}